\documentclass[11pt]{article}

\usepackage[margin=1in]{geometry}
\usepackage{amsmath,amssymb}
\usepackage{booktabs}
\usepackage{graphicx}
\usepackage{xcolor}
\usepackage[hidelinks]{hyperref}
\usepackage{enumitem}
\usepackage{caption}
\usepackage{pgfplots}
\pgfplotsset{compat=1.18}
\usetikzlibrary{shapes.geometric}

\title{A-Evolve-Training: Autonomous Post-Training of a 30B Model}

\author{%
  Zhan Shi \quad Bing He \quad Yisi Sang \quad Benoit Dumoulin \quad Hanqing Lu  \\
  Amazon \\
  \texttt{luhanqin@amazon.com}\\%
}
\date{}

\begin{document}
\maketitle

\begin{abstract}
Post-training a frontier model today is the work of a research team
iterating over weeks: proposing data-mix and recipe changes, launching
runs, reading evaluations, and deciding what to keep. We report a
\emph{autonomous post-training} system that runs this loop with no
human in the loop, performing post-training of a
30B-parameter Nemotron model across four rounds on GPU clusters over multiple weeks. The autonomously produced model reaches a held-out
score of $0.86$ against the top human submission's $0.87$ on the public
leaderboard of the NVIDIA Nemotron-Reasoning Challenge,\footnote{Public
leaderboard standing as of 2026-06-01, the date of this report.
Challenge URL: \url{https://www.kaggle.com/competitions/nvidia-nemotron-model-reasoning-challenge/overview}.} 
placing $8\text{th}$ of roughly $4000$ entries at the time of writing this report. Beyond
the headline number, the loop detected that its own internal development
metric had stopped tracking external performance on the visibly weakest
reasoning domain---candidates pushed it to record highs without moving
the external target---and revised its own search policy in response: it
stopped asking for higher dev and instead asked for interventions that
\emph{lowered} the now-misleading proxy while improving the external
target. \textbf{We treat this as direct, auditable evidence that a scaled
autonomous loop can produce \emph{discovery}, not only optimisation: the
loop did not merely optimise inside a fixed measurement frame; it
detected that the frame had become misleading and changed what counted
as evidence.} We adopt the operational view that any autonomous research system worth the ``recursive self-improvement'' label must eventually be able to perform end-to-end post-training of a
frontier-class model; the result reported here is one datapoint of
that bar being cleared. We deliberately avoid framing this as a ``first
autonomous match'' of human researchers. The claim we do make is narrower and auditable: to our knowledge, \textbf{this is the first
publicly reported autonomous post-training run at this scale} ---prior public
demonstrations of autonomous ML research operate at roughly GPT-2-class ($\sim$124M)
budgets.
As a separate scale-up, the same system has been applied to
post-train the \textbf{120B and 550B} Nemotron models end-to-end; with no public
human baseline at that scale in the NVIDIA Nemotron-Reasoning Challenge, this run evidences only that the autonomous loop \emph{closes} at 120B and 550B scale, not that its output is competitive with a human-authored recipe at that scale. We report it here as infrastructure evidence; the effectiveness claim is deferred until a comparable human anchor is available.
\end{abstract}

\section{Introduction}
\label{sec:intro}

Autonomous research systems are often discussed as a route toward
recursive self-improvement, but most public evidence today sits at a
scale that does not obviously transfer. Public demonstrations of
autonomous machine-learning research operate at roughly GPT-2-class
(\textasciitilde{}124M parameters) budgets~\cite{nanogpt_speedrun}:
small models, short runs, cheap evaluation loops. Systems
such as the AI Scientist line of work~\cite{ai_scientist} and related autonomous-research
agents show that the research loop \emph{closes} at that scale, but a
closed loop on a toy budget is not yet evidence that the loop closes
where the cost structure is an order of magnitude harsher. A useful
operational version of the long-term goal is concrete: an autonomous
research system worth the recursive-self-improvement label should
eventually be able to perform end-to-end post-training of a
frontier-class model. No prior public demonstration has cleared that
bar. We position this paper as one datapoint of that bar being
cleared---not a milestone, and at a scope ($30$B, multi-week campaign,
multi-H200-GPU clusters) precisely enough specified that the claim can be
audited rather than merely believed.

Frontier post-training is a particularly unforgiving testbed for this
question, and that is precisely why we chose it. We decompose a
research-iteration loop into four elements---\emph{hypothesis} (what to
try), \emph{execution} (running a single trial), \emph{strategy} (how
trials are allocated across a fixed budget), and \emph{infra} (the
substrate the other three run on)~\cite{aevolve_memo}---and ask how
each scales from GPT-2-class research to end-to-end frontier training.
The inflation is large and uneven (Table~\ref{tab:cost}): execution and
infra grow by orders of magnitude, while the hypothesis space widens
less in raw count than in kind. The deeper consequence is qualitative,
not merely quantitative. At 124M, the seed-level noise in any single run
can be averaged away with cheap repeats, so one run is a trustworthy
read and retrying, branching, and exhaustively sweeping are all free; at
30B, each repeat is a full training run, and the \emph{same} noise
becomes binding. The moves that are free at small scale are exactly the
ones that become prohibitive, so the binding constraint of the loop
extends beyond generating ideas to executing and measuring them.

\begin{table}[!htbp]
\centering
\small
\caption{Concrete per-element cost gap between GPT-2-class autonomous
ML research and frontier post-training. Each row gives the
mechanistic reason the element inflates with scale; the rough ratios
in the last column are illustrative order-of-magnitude estimates, not
measurements. The retry-cheaply assumption that closes the loop at
small scale becomes prohibitive once execution is a multi-week H200 training run
and infra is a research artifact in its own right.}
\label{tab:cost}
\renewcommand{\arraystretch}{1.25}
\begin{tabular}{@{}p{1.5cm} p{4.7cm} p{5.6cm} p{1.3cm}@{}}
\toprule
\textbf{Element} & \textbf{GPT-2-class ($\sim$124M)} & \textbf{Frontier post-training (30B, this work)} & \textbf{ratio} \\
\midrule
hypothesis
  & narrow dimension space: mostly architecture coefficients and optimiser hyperparameters
  & Wide design space: synthetic data construction, SFT/RL choice, loss design, schedules, data mixture, augmenters, checkpoint selection, evaluation design.
  & $\sim$10$\times$ \\
execution
  & Minutes per training run on a short, self-contained codebase. Autoresearch fixes each experiment to a 5-minute budget.
  & multi-week H200 training runs over a production training stack, data pipeline, checkpointing, vLLM evaluation.
  & $\sim$10$^3 \times$ \\
strategy
  & feedback arrives within minutes, so broad sweeping, retrying, and local hill-climbing are affordable.
  & feedback arrives per trial after hours/days, often noisy or distribution-shifted, so budget must be allocated carefully across a few hypotheses.
  & $\sim$10$^2 \times$ \\
infra
  & single off-the-shelf PyTorch script, single GPU, one metric, no distributed orchestration.
  & multi-H200-GPU Kubernetes orchestration, persistent storage, checkpoint management, evaluation harness, leaderboard submission, failure recovery.
  & $\sim$10$^2 \times$ \\
\bottomrule
\end{tabular}
\end{table}

We close part of this gap with a first-at-scale autonomous
post-training system. Three design choices, each shaped by the cost
structure above, make the loop survive contact with frontier-scale
execution. First, an \emph{immutable reference substrate}: every round
forks the same operator-audited default training stack into isolated
candidate sandboxes and never overwrites the substrate, which is what
keeps recipes comparable across rounds when each trial costs a
multi-week H200 training run. Second, \emph{homogeneous, memory-free workers}. We started with
what looks like the obvious design: specialised agents (a data
agent, a training agent, an eval agent) handing off mid-states like
a human research team. It did not scale. Compounding from
mid-states compounds \emph{unobserved variance} along with the
intended change and corrupts the very signal selection depends on.
The configuration that worked is the opposite---each round spawns
$N{=}8$ identical full-stack agents ({one as a pure-baseline
anchor for noise calibration, seven exploring orthogonal axes}) that
edit training recipes and data pipelines, launch GPU jobs, debug failures,
evaluate checkpoints, and write results, with no state carried across rounds. Third,
\emph{round-level evidence aggregation under a constitutionally
bounded meta agent}: a one-shot collector summarises cross-worker
evidence into a fixed-schema round report, and a meta agent rewrites
only the next round's rolling search policy under a frozen
constitution it cannot itself modify. The substrate is never
overwritten by a winning recipe; promotion lives entirely in the
rolling policy, not in the substrate. A
single cross-cutting principle---\emph{asymmetric freedom}---ties the
three together: zero degrees of freedom on the axes that must stay
invariant for trials to remain comparable, maximal freedom on the
axes where exploration creates value. We defer the mechanics to
Section~\ref{sec:system}, where Figure~\ref{fig:system} gives a
system overview.

Across four rounds, the autonomously
produced model reaches a score of $0.86$ on the challenge's public
leaderboard against the top score of $0.87$ at submission time, ranking
$8$th among roughly $4000$ entries. The score
is the headline datapoint, but not the main so-what. The system
answers \emph{how} a frontier-scale autonomous loop can be made to
close at this cost structure; what follows answers \emph{what} closing
it produced beyond the headline number. More interesting than the final score was a strategic reversal the loop discovered on its own. Early rounds treated the internal development metric as the natural object of optimisation: find the weakest-looking domain, add data or cleaner reasoning traces for it, and promote recipes that raise the measured score. This worked until it did not.

In later rounds, the loop found interventions that drove the internal metric to record highs, especially by fitting one visibly weak reasoning domain, but those gains failed to transfer to the external target. The lesson was not simply that one recipe failed. The loop had falsified a premise of its own search: the easiest proxy dimension to improve was no longer the causal bottleneck.

The next search policy therefore inverted the objective. Instead of asking for higher internal scores, it explicitly asked for interventions that might lower the proxy while improving the external target, such as rebalancing overrepresented domains and selecting checkpoints by a reweighted criterion. We view this as a stronger form of discovery than a one-off recipe improvement: the autonomous loop did not merely optimise within a fixed measurement frame; it detected that the frame itself had become misleading and changed what counted as evidence.

The remainder of the paper proceeds as follows.
Section~\ref{sec:related} situates the work relative to LLM-as-scientist
agents, LLM-guided evolutionary code search, and classical AutoML.
Section~\ref{sec:system} develops the system design under the cost
structure above, including the asymmetric-freedom principle.
Section~\ref{sec:experiments} reports the four-round trajectory.
Section~\ref{sec:discovery} details the proxy-reversal finding.
Section~\ref{sec:discussion} draws the two takeaways: what scaled
autonomous loops can produce, and a design lesson for building them.
The system reported here was developed using an internal version
of the A-Evolve framework~\cite{aevolve_framework}.

\clearpage
\section{Related work}
\label{sec:related}

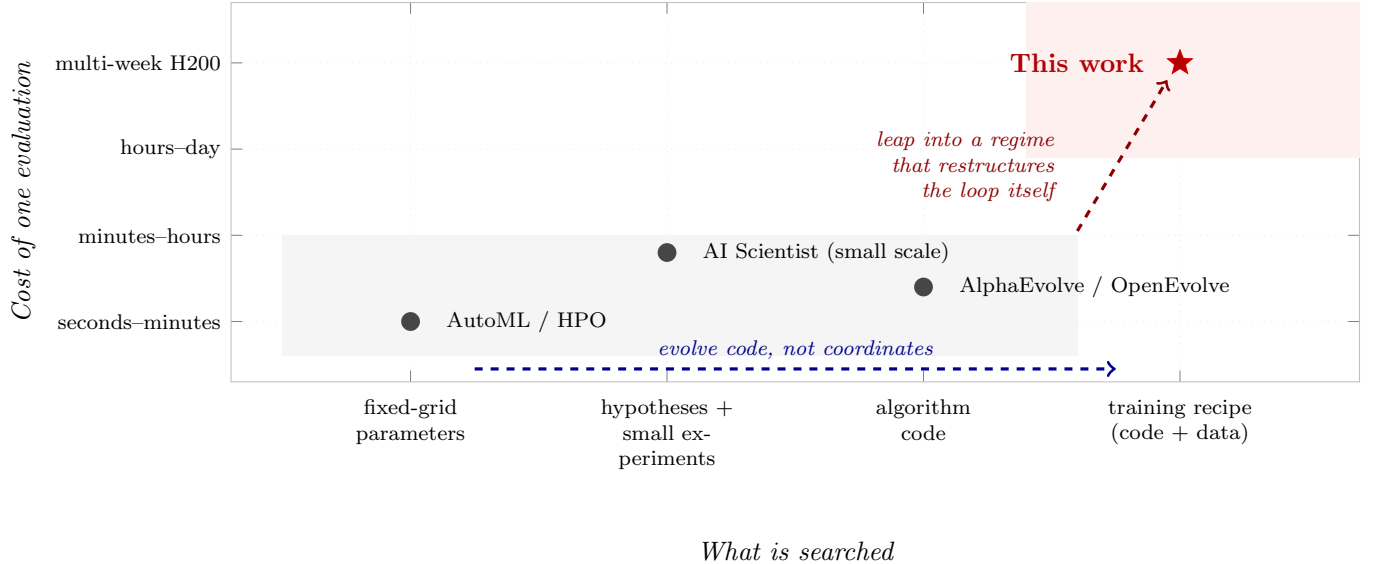
\begin{figure}[!t]
\centering
\begin{tikzpicture}
\begin{axis}[
    width=\linewidth, height=6.6cm,
    xlabel={\itshape What is searched},
    ylabel={\itshape Cost of one evaluation},
    xlabel style={font=\small, yshift=-7mm},
    ylabel style={font=\small},
    xmin=0.3, xmax=4.7,
    ymin=0.3, ymax=4.7,
    xtick={1,2,3,4},
    xticklabels={
      {fixed-grid\\parameters},
      {hypotheses +\\small experiments},
      {algorithm\\code},
      {training recipe\\(code + data)}
    },
    xticklabel style={align=center, font=\scriptsize, text width=2.2cm, yshift=-1mm},
    ytick={1,2,3,4},
    yticklabels={
      seconds--minutes,
      minutes--hours,
      hours--day,
      multi-week H200
    },
    yticklabel style={font=\scriptsize},
    grid=major, grid style={gray!15, dotted},
    axis line style={gray!50},
    clip=false,
]

\fill[red!6] (axis cs:3.4,2.9) rectangle (axis cs:4.7,4.7);
\fill[gray!8] (axis cs:0.5,0.6) rectangle (axis cs:3.6,2.0);

\draw[->, very thick, blue!55!black, dashed]
    (axis cs:1.25, 0.45) -- (axis cs:3.75, 0.45);
\node[font=\scriptsize\itshape, blue!55!black, anchor=south]
    at (axis cs:2.5, 0.45) {evolve code, not coordinates};

\draw[->, very thick, red!55!black, dashed]
    (axis cs:3.6, 2.05) -- (axis cs:3.95, 3.8);
\node[font=\scriptsize\itshape, red!55!black, anchor=east, align=right]
    at (axis cs:3.55, 2.8) {leap into a regime\\that restructures\\the loop itself};

\node[circle, fill=gray!55!black, inner sep=2.5pt] at (axis cs:1, 1) {};
\node[font=\scriptsize, anchor=west] at (axis cs:1.1, 1) {AutoML / HPO};

\node[circle, fill=gray!55!black, inner sep=2.5pt] at (axis cs:2, 1.8) {};
\node[font=\scriptsize, anchor=west] at (axis cs:2.1, 1.8) {AI Scientist (small scale)};

\node[circle, fill=gray!55!black, inner sep=2.5pt] at (axis cs:3, 1.4) {};
\node[font=\scriptsize, anchor=west] at (axis cs:3.1, 1.4) {AlphaEvolve / OpenEvolve};

\node[star, star points=5, star point ratio=2.3,
      fill=red!70!black, draw=red!85!black, inner sep=1.5pt]
    at (axis cs:4, 4) {};
\node[font=\small\bfseries, color=red!70!black, anchor=east]
    at (axis cs:3.9, 4) {This work};

\end{axis}
\end{tikzpicture}
\caption{Positioning of this work against three related lines along
two principal axes: \emph{what is searched} (horizontal, from
fixed-grid parameters to a training recipe of code and data) and
\emph{the cost of one evaluation} (vertical, from seconds-to-minutes
to multi-week H200). On the horizontal axis, AlphaEvolve and our
system stand together against classical AutoML---evolve code, not
coordinates. On the vertical axis, AutoML / HPO, AI Scientist, and
AlphaEvolve all sit in the low-eval-cost cluster (per-trial
evaluation is comparable to or cheaper than a small training run);
this work occupies the multi-week per-trial regime alone, and the
discontinuity is what forces the loop itself to be restructured
around the cost. Positioning is approximate; finer-grained
variation within each family exists.}
\label{fig:positioning}
\end{figure}

\paragraph{LLM-as-scientist at small scale.}
The AI Scientist line of work~\cite{ai_scientist} and related
autonomous-research agents demonstrate that an LLM-driven loop can form
hypotheses, run experiments, and write up findings with no human in the
loop. This class of work operates at budgets where an ``experiment'' is
cheap, the model under study is small, and a failed run can be retried
freely; its canonical substrate is GPT-2-scale training on Karpathy's
nanoGPT~\cite{nanogpt}. Even in that forgiving regime the problem is far
from solved: on the nanoGPT speedrun, recent evaluation
work~\cite{nanogpt_speedrun} reports that frontier agents struggle to
reimplement \emph{known} improvements at {GPT-2 scale}, even when
handed detailed hints. Our contribution is orthogonal to both the
demonstrations and these evaluations, along the axis that matters most
here---\emph{scale}. We do not claim a better agent; we report what
changes when each trial in the loop is an end-to-end post-training
run on a 30B model, so that the cheap-retry assumption no longer holds
and the loop must nevertheless converge.

\paragraph{LLM-guided evolutionary code search.}
A second line of work uses LLMs to drive evolutionary search directly
over program code rather than over fixed parameter coordinates.
AlphaEvolve~\cite{alphaevolve} evolves candidate algorithms (scheduling
primitives, mathematical procedures, hardware-level code) under an LLM
mutation operator, and reports new results on a range of open algorithmic
problems, including a faster matrix-multiplication algorithm. Open
replications such as OpenEvolve~\cite{openevolve} have made this style of
agent broadly available. The object under search in these systems is an
\emph{algorithm}: the evolver edits a program, and an automated,
machine-gradable evaluator scores whether its behaviour is better. The
object under search in our system is a \emph{training recipe}: the
evolver edits the code and data that produce a post-trained model, and
the evaluator measures the downstream model's behaviour. Both share the
LLM-driven mutation primitive; the feedback structure and the cost of
each evaluation differ by orders of magnitude.

\paragraph{Classical AutoML and hyperparameter search.}
Classical AutoML and hyperparameter optimisation
\cite{automl_survey} search a parameterised space---learning rates,
schedules, architecture coefficients---under a fixed pipeline. Our
loop evolves a different object: the \emph{code and data} of the
training recipe itself (the data-construction pipeline, training data mixture
weights, and the trainer configuration), not coordinates in a
pre-declared hyperparameter space. The distinction is not cosmetic:
the upsample finding reported in Appendix~\ref{app:rounds} (Table~\ref{tab:rounds}, Round~3) is a change to
the \emph{data-mixture program}, a move a hyperparameter search over
a fixed grid could not express.

\paragraph{Where this work sits.} Figure~\ref{fig:positioning}
places the four lines along two principal axes---\emph{what is
searched} and \emph{the cost of one evaluation}---and locates this
paper alone in the multi-week per-trial regime where the
surrounding loop itself must be restructured around the cost.

\section{System design}
\label{sec:system}

\begin{figure}[!htbp]
\centering
\includegraphics[width=0.95\linewidth]{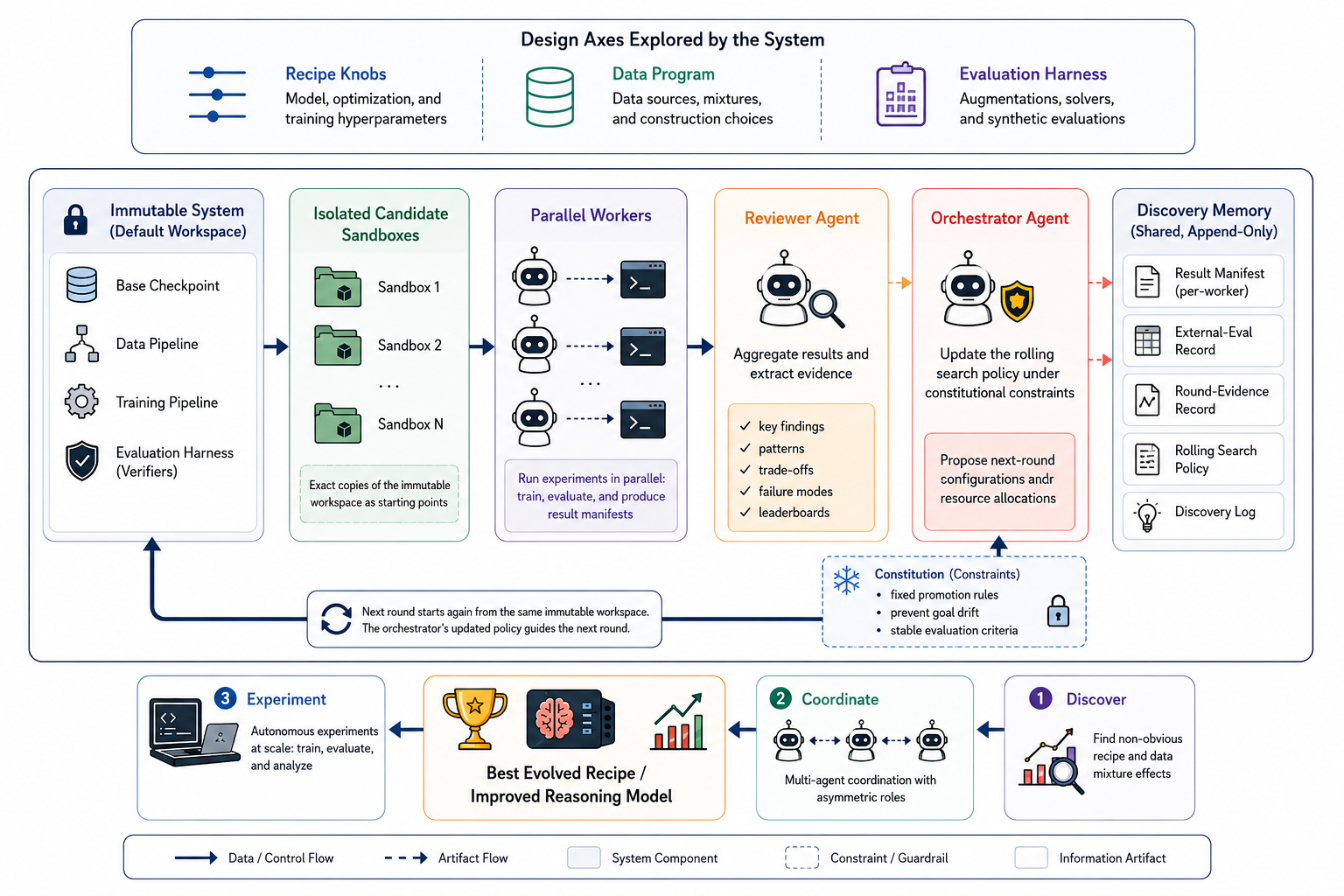}
\caption{System overview. An immutable operator-audited substrate
(\texttt{default/}) is forked into $N$ candidate sandboxes per round.
Memory-free workers explore an axis assigned by the current round's
research policy, dispatch real GPU training jobs, evaluate at
multiple checkpoints, and write a structured \emph{results summary}. A
reviewer cross-reads worker outputs into a fixed-schema summary; a
constitutionally bounded orchestrator updates only the next round's
research policy under a frozen meta-prompt. The default substrate is
never overwritten, and cross-round state reduces to the rolling
research policy and a consolidated discovery log.}
\label{fig:system}
\end{figure}

We describe the system in terms of \emph{why it must be shaped this way
to scale}, rather than as a list of components. The governing constraint
is the one from Section~\ref{sec:intro}: once a single trial is a
multi-week training run, any design that multiplies trials, that lets
trials drift out of comparability, or that requires a fragile pipeline to
be regenerated per trial, does not survive contact with the cost
structure. Figure~\ref{fig:system} shows the resulting system at a
glance; the three pillars and the cross-cutting principle below
specify what each component must do, and why.

\paragraph{Pillar 1: an immutable reference substrate.}
The system is anchored on a single human-verified workspace (named as \texttt{default/} here),
 containing a base checkpoint, a data-construction pipeline,
model training and evaluation code. Every round forks
from \texttt{default/}; crucially, the substrate is \emph{never}
overwritten by a winning recipe, and no round forks from a previous
round's winner. This is a deliberate ``constraint folding'' move: the
human one-shot pre-solves everything that must be correct for a run to
execute at all (APIs, data formats, metric alignment, cluster config),
leaving the workers a space of semantically meaningful variation rather
than a space in which most points fail to run. Re-forking from a fixed
substrate is also what keeps recipes comparable across rounds---a recipe
measured in round~4 is measured against the same ground as one from
round~1.

\paragraph{Pillar 2: homogeneous, memory-free workers.}
Each round launches $N$ identical worker sessions that differ only in
which axis each is asked to explore (a data-mix change, a schedule
change, a filtering change, and so on). There are no specialised
roles---no ``data agent'' handing mid-state outputs to a ``trainer
agent''---and no memory is carried across rounds. We adopted this only
after the opposite design failed: an earlier version with specialised
roles iterating on each other's intermediate outputs, built to mirror a
human research team, did not work in practice. We attribute the inversion
to two structural facts about current-generation coding agents. First,
cross-axis reasoning---not bandwidth---is the scarce resource, so role
specialisation fragments the one capability that is actually limiting and
destroys more value than it creates. Second, compounding from mid-states
compounds \emph{unobserved variance} along with the intended change,
which breaks recipe-comparability across rounds and corrupts the very
signal selection depends on.

\paragraph{Pillar 3: round-level evidence aggregation and policy-only promotion.}
Selection runs at the granularity of a round, but with a deliberate
constraint: the substrate ({\texttt{default/}}) is never overwritten,
even by a winning recipe. Promotion therefore happens at the
\emph{policy} level rather than at the \emph{substrate} level. The
round closes through three structured artifacts and one read-only
constitution:

\begin{itemize}[leftmargin=14pt, itemsep=1pt, topsep=1pt]
\item {Per-candidate \emph{result manifest}}: each of the
{$N{=}8$ candidate workers} emits a typed manifest recording the
axis explored, the recipe diff applied, training and per-checkpoint
evaluation scores, and observed failure modes; an append-only
{execution trace} carries the worker's wall-clock progress and is
used as the recovery source on candidate crash.

\item Round-evidence artifact: a one-shot collector agent
ingests every candidate's result manifest, fetches the public
leaderboard, and emits a {fixed five-section record}---{per-candidate
summaries, cross-candidate patterns, dev/leaderboard calibration,
surprises, and a design-rationale audit}. The schema is invariant
across the campaign; what mutates round-to-round is the content inside
each section, not the section list itself.

\item {Rolling search policy}: the meta agent rewrites the next
round's search policy from the round-evidence artifact under a
{system-level constitution} the agent has {read-only access to
by construction} (the constitution is loaded as the agent's frozen
system prompt; the agent has no write handle to it).
\end{itemize} The meta agent (i) updates the
{standing-recipe specification} (the working hyperparameter set
subsequent candidates inherit on their fork), (ii) promotes or
retires {search axes} under the constitution's meta-rules
{(leaderboard authoritative; dev/leaderboard gap as a calibration
signal; axis-category rotation when leaderboard improvement
plateaus across two rounds)}, and (iii) maintains a monotonically
growing {dead-end registry} (literal section title in the rolling
policy: {``Don't waste budget on''}) with round-tagged provenance
on each entry. Any empirical noise rules---including the current
promotion threshold itself, calibrated to the observed seed-variance
band---live in the {rolling search policy} and are themselves
discoverable and revisable across rounds. The
constitution enforces the meta-rules; the rolling policy carries
the empirical rules that arise from data. Because the post-training
feedback is drawn from the \emph{same distribution} as the objective
(we measure the quantity we want, only varying the recipe), a
recipe's measured score is a usable predictor of the next recipe's
score under either layer.

\paragraph{Cross-cutting principle: asymmetric freedom.}
The three pillars are unified by a single principle: freedom is granted
\emph{asymmetrically} across the design's two axes. Along the axis that
must stay invariant for trials to remain runnable and comparable---the
substrate, the evaluation harness, the comparison baseline---workers have
\emph{zero} freedom. Along the axis where exploration creates value---which
semantic mutation to attempt---workers have \emph{maximal} freedom. The
system's ability to scale comes precisely from this asymmetry. The failed
specialised-role design failed because it loosened the invariant axis
(letting mid-states compound and drift); the working design succeeds
because it freezes that axis hard and spends all of its degrees of
freedom on the axis that pays.

\section{Results}
\label{sec:experiments}
We tested our A-Evolve-Training system on the NVIDIA Nemotron-Reasoning Challenge. In this competition, participants work from a shared Nemotron 3 Nano baseline and a novel reasoning benchmark developed by NVIDIA Research. Nemotron provides an open foundation for this challenge, datasets, and training recipes that participants can build on or adapt within their own workflows. This dataset comprises a collection of logical reasoning puzzles requiring the identification and application of underlying transformation rules. The puzzles cover various domains, such as bit manipulation and algebraic equations.

\begin{figure}[!htbp]
\centering
\begin{tikzpicture}
\begin{axis}[
    width=0.8\linewidth, height=5.2cm,
    xlabel={Round}, ylabel={Leaderboard score},
    xmin=0, xmax=4, ymin=0.78, ymax=0.88,
    xtick={0,1,2,3,4}, ytick={0.78,0.80,...,0.88},
    legend pos=south east, legend style={font=\small},
    grid=both, grid style={gray!20},
]
\addplot[thick, mark=*] coordinates {(0,0.80)(1,0.82)(2,0.84)(3,0.85)(4,0.86)};
\addlegendentry{autonomous loop}
\addplot[dashed, thick] coordinates {(0,0.87)(4,0.87)};
\addlegendentry{human \#1 ($0.87$)}
\end{axis}
\end{tikzpicture}
\caption{Leaderboard score by round. The autonomous loop closes most of
the gap to the top human submission over four rounds and finishes one
point ($0.86$ vs.\ $0.87$) below it, at rank $8/{\approx}4000$.}
\label{fig:trajectory}
\end{figure}
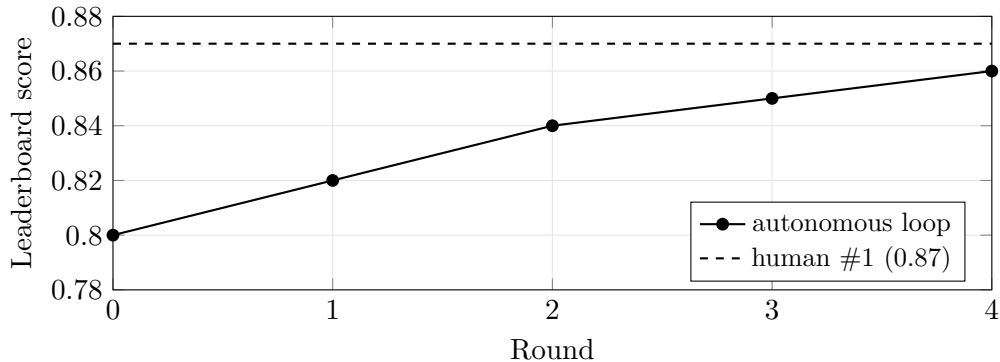

We ran the autonomous campaign for {four rounds} on a {30B Nemotron
base}~\cite{nvidia_nemotron_nano_v3_2025} with {$N{=}8$ workers per round} and no human intervention
between the initial substrate authoring and the final submission. The
loop improved monotonically across rounds (Figure~\ref{fig:trajectory})
and converged to a held-out leaderboard score of {$0.86$} against
the top human submission's $0.87$. The full round-by-round trajectory---each
round's explored axes and the load-bearing findings the meta agent
extracted from them---is reported in
Appendix~\ref{app:rounds}.


\section{Discovery: when the proxy stopped being evidence}
\label{sec:discovery}

We single out the loop's \textbf{strategic reversal} because it is stronger
evidence for autonomous discovery than any single recipe improvement.
Early rounds treated the internal development metric as the natural object
of optimisation: identify the weakest-looking domain, add data or cleaner
reasoning traces for it, and promote recipes that raise the measured
score. This assumption was initially useful. But in later rounds the loop
found interventions that drove the internal metric to record highs while
failing to move the external target. In the clearest case, candidates
pushed overall dev to {$0.93$--$0.94$} and lifted the visibly weak
equation domain from a long-standing {${\sim}0.65$ ceiling to as high
as $0.82$}. Under the ordinary proxy-optimisation story, this should have
been a breakthrough. It was not: external performance remained in the
same {$0.84$--$0.85$} band.

The lesson was not simply that one recipe failed. The loop had falsified
a premise of its own search: the easiest proxy dimension to improve was
no longer the causal bottleneck. A follow-up intervention made this more
explicit. The loop downweighted the same domain that had looked like the
obvious weakness; the internal score on that domain fell, but the external
target did not degrade. Together, the positive and negative interventions
showed that the domain attracting the most optimisation pressure under
the proxy was not the limiting factor for real performance.

This changed the search policy itself. The next round no longer asked
workers to maximise raw dev. It asked for interventions that could
\emph{lower} the proxy while improving the external target: rebalancing
overrepresented domains, selecting checkpoints by a reweighted criterion,
and treating raw dev gains as suspect unless they survived an external
probe. We view this as discovery at the level of the research loop. The
system did not merely optimise inside a fixed measurement frame; it
detected that the frame had become misleading and changed what counted as
evidence.

\paragraph{What this observation rests on.}
The reversal rests on multiple single-variable probes rather than on a
single lucky run: high-dev synthetic interventions that failed to transfer,
and a downweighting intervention that damaged the apparent weak domain
without damaging the external target. The conclusion is still scoped: it
does not prove that the internal metric is useless, only that beyond this
performance band it ceased to be a reliable causal guide. That distinction
is exactly the point. At frontier scale, discovering \emph{when not to
trust the proxy} is itself part of the research problem.

\section{Discussion}
\label{sec:discussion}

\subsection{Scaled autonomous loops can revise the rules of their own
search, not only run within them}
\label{sec:disc-sowhat}

The leaderboard result shows the loop is a competent optimiser. The
proxy-reversal finding (\S\ref{sec:discovery}) shows something we think
is more important: a scaled autonomous loop can revise the rules of its
own search. Optimisation tightens what the operator already believes
is worth searching; the loop here did not just search harder---it
detected that the measurement frame on its visibly weakest domain had
stopped being evidence, and changed what it asked workers to maximise.
We are deliberately narrow about the strength of the claim---it rests,
so far, on a single un-replicated campaign---but the direction is the
point. If even one decision in four rounds lands outside the operator's
prior \emph{about what counts as evidence} and pays, the expected yield
of \emph{discovery} (as distinct from optimisation) grows with the
number of autonomous trials the loop can afford, which is exactly the
quantity scale buys. The auditable version of the claim is therefore:
at this scale, the loop produced at least one verifiable revision of
its own search policy that a within-prior optimiser is not built to
produce.

\subsection{A system-design lesson: asymmetric freedom and memory-free
workers}
\label{sec:disc-design}

The second takeaway is a design principle rather than a headline. The
working configuration freezes the invariant axis hard---one immutable
substrate, one fixed evaluation, memory-free workers that re-fork rather
than carry state---and spends all of its freedom on the axis where
variation creates value. Memory-free re-forking is not a limitation we
tolerated; it is load-bearing, because it is what preserves
cross-round comparability and prevents the unobserved-variance compounding
that sank the specialised-role design. The transferable lesson for anyone
building an autonomous training loop is that the instinct to mirror a
human research team---specialised roles, shared evolving state, agents
building on each other's mid-states---inverts the right design. Constrain
the agents on what must stay invariant; free them only on what pays.

The same principle operates a level up, across rounds rather than
within them. The substrate (\texttt{default/}) and the constitution
(the meta agent's frozen system prompt) are immutable across the
entire campaign; the rolling search policy is the only artifact that
mutates, and it does so under rules the constitution sets but does
not itself express. The constitution encodes the meta-rules
(leaderboard authoritative, axis-category rotation on plateau,
dev/leaderboard-gap calibration); the rolling policy carries the
empirical rules the system discovers from data (current promotion
threshold, standing recipe, which axes have retired). This
two-level split is what lets the loop stay disciplined---no goal
drift, no substrate contamination across rounds---and learn---revise
its own noise band, retire entire axis categories---at the same
time. It is a second instance of asymmetric freedom: freeze the
invariant axis, free the axis where variation creates value, then
iterate the layer above under the same constraint.

\subsection{Mapping back to the operational RSI framing}
\label{sec:disc-rsi}

Returning to the framing of \S\ref{sec:intro}: we adopted the
operational view that any autonomous research system worth the
recursive-self-improvement label should eventually be able to perform
end-to-end post-training of a frontier-class model. The result reported
here clears that bar once, narrowly defined---a 30B Nemotron base, four
self-directed rounds, no human in the loop, finishing within one point
of the top human submission. We do not read this as evidence that
recursive self-improvement is solved or close to solved; the dependence on a single base model
and a single benchmark, and the pre-audited substrate that bounds
the search space all remain real limits, and each is the subject of an
entry in Section~\ref{sec:future}. What we do read it as is evidence
that the bar---autonomous frontier post-training---is in fact
reachable, and that the loop, once reachable, can revise its own
evidence rules in ways the operator would not have prescribed. The interesting next
question is no longer whether the bar can be cleared at this scale,
but how reliably it can be cleared, on what tasks, and under which
system designs.

\section{Future work and limitations}
\label{sec:future}

\begin{itemize}[leftmargin=1.4em, itemsep=2pt]
  \item \textbf{More domains.} The result is on a single post-training
    task family; whether the loop produces out-of-prior findings on other
    domains is open.
  \item \textbf{More base models.} We ran one 30B Nemotron base.
    Replication on other bases and sizes would test whether the system's
    behaviour is a property of the loop or of this checkpoint.
  \item \textbf{More benchmarks.} We rely on one public leaderboard as the
    external anchor; a broader benchmark would strengthen the
    optimisation claim and reduce dependence on a single metric.
  \item \textbf{Code release.} The system reported here was developed
    using an internal version of the A-Evolve framework~\cite{aevolve_framework};
    full code, reference substrate, and trained model checkpoint
    release timing is to be confirmed.
\end{itemize}


\clearpage
\appendix
\section{Round-by-round trajectory}
\label{app:rounds}

\begin{table}[!htbp]
\centering
\caption{Round-by-round trajectory of the autonomous campaign. Scores
mirror Figure~\ref{fig:trajectory}. The central discovery is a strategic
one: the loop first optimises the internal proxy, then discovers where
that proxy stops tracking the external target, and finally changes the
search policy accordingly.}
\label{tab:rounds}
\renewcommand{\arraystretch}{1.25}
\small
\begin{tabular}{@{}p{1.2cm} p{3.2cm} c c p{6.8cm}@{}}
\toprule
\textbf{Round} & \textbf{Search mode} & \textbf{Dev} & \textbf{LB} & \textbf{What the loop learned} \\
\midrule
0
& Substrate
& 0.79
& 0.80
& Fixed human-authored baseline. All candidates fork from this substrate;
winning recipes update the search policy, not the substrate itself. \\

1
& Recipe stabilisation
& 0.82
& 0.82
& Basic training levers were still causal: schedule shape, optimizer
hygiene, LoRA capacity, and clipping moved both the proxy and the external
target. At this stage, maximising dev was a reasonable strategy. \\

2
& Stack-and-calibrate
& 0.84
& 0.84
& Early improvements stacked, but seed noise became large enough that
small dev gains could no longer be trusted. The loop began treating
replication and calibration as part of the research problem. \\

3
& Proxy exploitation
& 0.93
& 0.85
& The loop found how to drive dev to record highs, especially by fitting
the visibly weak equation domain from a long-standing ${\sim}0.65$ ceiling
to as high as $0.82$. But the external target barely moved.
This falsified the premise that the easiest proxy dimension to improve was
the causal bottleneck. \\

4
& Proxy-aware search
& 0.89
& 0.86
& The loop changed strategy: instead of simply raising dev, it searched for
interventions that should transfer under the external distribution, such
as domain rebalancing and reweighted checkpoint selection. This recovered
external progress and produced the campaign-best LB score. \\

\bottomrule
\end{tabular}
\end{table}

\end{document}